# Deep Learning for automated multi-scale functional field boundaries extraction using multi-date Sentinel-2 and PlanetScope imagery: Case Study of Netherlands and Pakistan

**Saba Zahid**
Space Science department
Institute of Space Technology
Islamabad, Pakistan
sabazahid.rsgis@gmail.com

*Sajid Ghuffar*
*Space Science department*
*Institute of Space Technology*
*Islamabad, Pakistan*
sajid.ghuffar@grel.ist.edu.pk

*Obaid-ur-Rehman*
*Space Science department*
*Institute of Space Technology*
*Islamabad, Pakistan*
obaidurrehman@grel.ist.edu.pk

**Syed Roshaan Ali Shah**
Independent Researcher,
Pakistan

roshaan9@yahoo.com

*Abstract*— Monitoring crop health, ensuring food security, and implementing precision agriculture depend on accurate knowledge of agricultural field boundaries. Rapid temporal changes in agriculture field make challenge to abstract information from single date imagery in crop growing season. This task also become difficult in areas with small landholdings and lack of ground-based data for training of deep learning models. This study explores the effectiveness of multi-temporal satellite imagery for better field boundary delineation using deep learning semantic segmentation architecture on two distinct geographical and multi-scale farming systems of Netherlands and Pakistan. Multidate images of April, August and October 2022 were acquired for PlanetScope and Sentinel-2 in sub regions of Netherlands and November 2022, February and March 2023 for selected area of Dunyapur in Pakistan. For Netherlands, Basic registration crop parcels (BRP) vector layer was used as labeled training data. while self-crafted field boundary vector data were utilized for Pakistan. Four deep learning models with UNET architecture were evaluated using different combinations of multi-date images and NDVI stacks in the Netherlands subregions. A comparative analysis of IoU scores assessed the effectiveness of the proposed multi-date NDVI stack approach. These findings were then applied for transfer learning, using pre-trained models from the Netherlands on the selected area in Pakistan. Additionally, separate models were trained using self-crafted field boundary data for Pakistan, and combined models were developed using data from both the Netherlands and Pakistan. Results indicate that multi-date NDVI stacks provide additional temporal context, reflecting crop growth over different times of the season. This helps deep learning models learn the aggregated vegetation response of crop fields as functional units, enhancing model performance. The study underscores the critical role of multi-scale ground information from diverse geographical areas in developing robust and universally applicable models for field boundary delineation. Incorporating geographical diversity and comprehensive ground data, alongside temporal context from multi-date satellite data, is essential for improving the accuracy and applicability of predictive models, especially in regions with fragmented landholdings and limited ground-truth data. The results also highlight the importance of fine spatial resolution for extraction of field boundaries in regions with small scale framing. The findings can be extended to multi-scale implementations for improved automatic field boundary delineation in heterogeneous agricultural environments.

## I. Introduction

Agriculture is a dynamic subject characterized by incessant farm activities, diverse cropping patterns, and cyclic crop growth in both space and time, presenting challenges for autonomous farming decisions. Continuous monitoring of crop development is essential to account for spatial and temporal changes. Field boundaries play a fundamental role in crop management as they define the basic units of farm operations [1]. These boundaries are crucial for various aspects, including differentiating farm ownership, selecting crops, and effectively managing time. By providing reliable information, field boundaries enable informed decisions at the field level, facilitating crop monitoring and yield prediction [1].

Research on field boundary delineation primarily focuses on cadastral and functional field boundaries. Cadastral field boundaries pertain to official revenue records and primarily address land ownership [2]). In contrast, functional field boundaries delimit active crop fields and exhibit unique spectral and spatial patterns during the growing seasons. These boundaries are more effective than cadastral or other mapping methods for carrying out agricultural operations, particularly in precision agriculture and managing multiple crops. Functional field boundaries have various applications, including government agencies tracking incentives and food production [3], crop insurance under changing climate conditions with high climate hazard probability [4], mapping crop types [5], providing digital farm services [6], monitoring the spread of pests and diseases [7], and sowing different crop species [8]. They serve as a crucial tool for implementing precision agriculture.

However, delineating field boundaries poses challenges due to their non-trivial, irregular, and complicated features [9]. Traditional methods such as field surveys and manual digitization of aerial or satellite images are labor-intensive, costly, and time-consuming, making them impractical for covering large agricultural lands [10]. To address this, researchers have explored automatic field boundary delineation as a scalable solution [11] [12]. In addition to traditional techniques based on edge detection or object-based segmentation [13], [11], [14], [12]deep learning has emerged as a promising approach for functional field boundary delineation. Numerous studies have described the use of deep learning methods in this context [15], [16], [17]. The latest generation of algorithms for detecting (semantic) contours relies on deep learning networks. These networks have demonstrated impressive proficiency in acquiring meaningful data representations for tasks such as object recognition, image classification, and semantic segmentation [18]. Among various types of deep networks, Convolutional Neural Networks (CNNs) have gained significant popularity in image analysis [19]. These networks possess the ability to learn a hierarchical structure of spatial features across different layers, progressing from raw pixel values to object parts (boundaries), local shapes, contextual cues, and ultimately complex textural patterns [19].

Efforts have been made to increase the accuracy of field boundary delineation. For example, [20] delineated field boundaries using single-date Very High-Resolution WorldView-2 imagery in small regions of Kofa and Mali using Fully Convolutional Networks (FCN). In the context of smallholder systems, [17]used single-date Airbus SPOT and PlanetScope imagery for field boundary delineation in France, followed by prediction and improvement in regions of India using transfer learning. In absence of high-resolution images, [21] delineated field boundaries in Flevoland, the Netherlands using single date Sentinel-2 data by enhancing spatial context by up sampling in network layers novel super-resolution semantic contour detection network.

While advanced deep learning techniques have been implemented for field boundaries [16], delineation of automatic field boundaries across a range of agricultural environments remained a complex challenge [15]. The visual appearance of field edges becomes dynamic in space and time due to rapid temporal changes in agriculture fields caused by different crop types, crop rotations, sowing times, and variable crop growth. On a single-date image, static field boundaries may be more visible than functional field edges [10]. Various studies have extracted dynamically accumulating field boundaries over the growing season using multi-temporal Sentinel-2 images through traditional edge detection and image segmentation techniques [22] and multi-date aggregated vegetation index in regions of Russia and Ukrine [10]. Deep learning algorithms can leverage satellite image time series (SITS) to create more comprehensive and accurate delineations of functional field boundaries [15]. Deep learning, when combined with temporal stacking of multi-date satellite imagery, has shown improved field boundary delineation compared to single-date models, as demonstrated in Denmark [23].

To implement deep learning methods for field boundaries, ground label data is imperative for training the models. Vector-based annual crop layers, openly available for some European Union countries through the Land Parcel Identification System (LPIS), provide field-level digital crop information for subsidy validation claims [24]. These layers serve as a valuable basis for generating training data for field boundaries and have been utilized in various studies in countries such as France [25], Netherlands [21], and Spain [24]. However, this becomes a challenging task in many smallholder farming systems like India and Pakistan due to unavailability of labeled field boundaries data for training and validation of deep neural networks. This limitation can be resolved with manual creation of field boundaries data using visual interpretation on satellite imagery and annotations. Another approach may be applied as transfer learning from pre-trained deep learning model on the areas trained with ground labeled field boundaries data and first-hand predictions on small holder farmlands can further be used for further training after manual refinement [1]

This study serves as a proof of concept for using multi-date satellite data, particularly temporal NDVI stack, in semantic segmentation deep learning models to enhance the spatial context and temporal function of agriculture field boundary delineation in selected areas of the Netherlands and Pakistan. The study utilizes PlanetScope and Sentinel-2 multi-date images in conjunction with the Netherlands' basic registration crop parcels (BRP) vector layer as ground labels for field boundaries. Due to lack of field boundaries data in Pakistan, labeled vector data for field boundaries was indigenously digitized using visual interpretation on multi-date NDVI composite image.

## II. MATERIALS AND METHODS

### A. Study Area

Pakistan: In Pakistan Dunyapur tehsil was selected. This area belongs to major fertile agriculture plane with diversity of crops and field sizes. Total area of Tehsil is 995 km2. Due to limited time and manpower random sites were selected keeping in view the visual diversity of farm sizes. 74 sq km area was sub sampled for manual digitization and annotation of field boundaries which was further used in training of Deep learning models.

Netherlands: Ten randomly selected sites in Netherlands were chosen to cater for different agricultural / environmental conditions, cropping patterns and field size variability. These include areas in Drenthe (01 site) Overijssel (04 sites), Flevoland (01 Site) Noord-Holland (02 sites) and Noord-Branant (01 site) covering total area of 1790 km2 "Fig 1". These sub samples were taken keeping in view the limited quota of PlanetScope Imagery, as acquisition was carried out under educational and research program having limited data access from Planet.

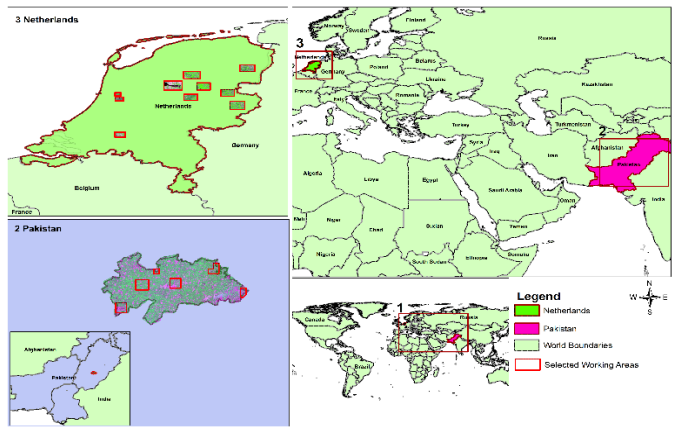

FIGURE 1: AREA OF STUDY; LOCATION OF SUBSETS AREAS IN NETHERLAND AND PAKISTAN

## B. Crop calender

The crop calendar is essential for optimizing image acquisition that enhance the accuracy and efficiency of deep learning models in detecting field boundaries. Aligning image acquisition with specific crop growth stages ensures clear visibility of field boundaries, as early or late stages may obscure boundaries due to bare soil or overlapping canopies. Further acquiring images during periods with minimal cloud cover and optimal vegetation conditions improves the clarity and quality of the data. In this study keeping in view limited computational resources long time series data was not used. To use multi-date images for deep learning, cropping pattern was considered for Pakistan (Fig 2i) and Netherlands (Fig 2ii) areas for identification of selective dates where appropriate crop response in space and time can be achieved to minimize the volume of data for computation without compromising the loss of important information. For Netherlands three distinct phases of agriculture are during April, August and October were observed and respectively data was acquired. For Pakistan, image acquisition was carried out for wheat season at early crop stage (November), mid vegetation stage (February) and Late crop stage (March).

| Crop Clander, Punjab, Pakistan | | | | | | | | | | | | |
|---|---|---|---|---|---|---|---|---|---|---|---|---|
| Crops | Jan | Feb | Mar | Apr | May | Jun | Jul | Aug | Sep | Oct | Nov | Dec |
| Wheat | | | | | | | | | | | | |
| Rice | | | | | | | | | | | | |
| Cotton | | | | | | | | | | | | |
| Sugarcane | | | | | | | | | | | | |
| Maize (Autumn) | | | | | | | | | | | | |
| Maize (Spring) | | | | | | | | | | | | |
| Key: | Sowing | | Growth | | Harvesting | | | | P: Picking | | T: Transplation | |

Source: Punjab Crop Reporting Services, Pakistan

| Crop Clander, Netherlands | | | | | | | | | | | | |
|---|---|---|---|---|---|---|---|---|---|---|---|---|
| Crops | Jan | Feb | Mar | Apr | May | Jun | Jul | Aug | Sep | Oct | Nov | Dec |
| Barley (Spring) | | | | | | | | | | | | |
| Barley (Winter) | | | | | | | | | | | | |
| Corn | | | | | | | | | | | | |
| Wheat (Spring) | | | | | | | | | | | | |
| Wheat (Winter) | | | | | | | | | | | | |
| Key: | | Plant | | Mid-Season | | Harvesting | | | | | | |

Source: FAS, USDA

FIGURE 2: CROP CALENDERS OF I) PAKISTAN II) NETHERLANDS

## C. Data Collection

To evaluate the use of multi-date images for field boundaries delineation using deep learning, two satellite data sources were used.

### 1) Sentinel 2 satellite imagery

Sentinel-2 consists of two counterpart satellites of S2-A and S2-B from European space Agency. Sentinel-2 provides open source optical imagery with 5-day revisit time. This frequency of data availability makes it important for high resolution time series remote sensing. Senitnel-2 provides 13 bands multi-spectral image ranging from 10m to 60m spatial resolution. However, for this study 04 spectral band Red, Green, Blue and NIR with native 10m resolution were used. Google earth engine was used to synthesize multi-date median images for three months i.e. for selected areas in Netherlands April, August and October 2022 and selected area of Pakistan November 2022, February and March 2023. Sentinel-2 harmonized product (COPERNICUS/S SR_ HARMONIZED) was downloaded for each date. The assets contain 12 UINT16 spectral bands representing SR scaled by 10000. NDVI was calculated for each of selected month image and temporally stacked to get 3 date NDVI composite image "Fig 3". Imagery was acquired for randomly selected subset areas to include maximum possible heterogeneity in training dataset.

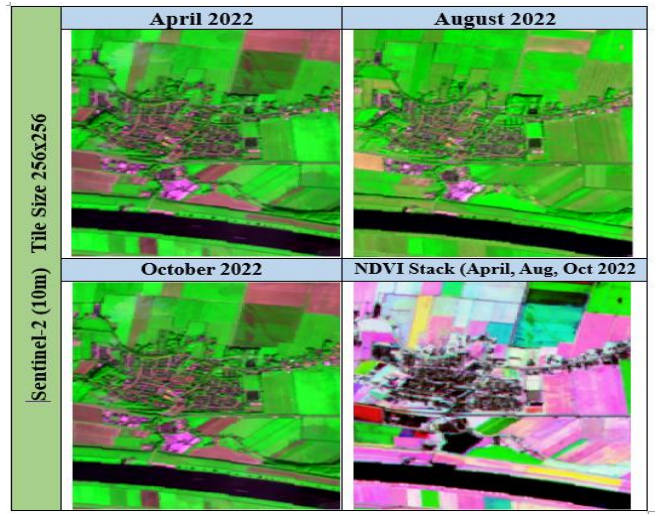

FIGURE 3: SENTINEL-2, MULTI DATE IMAGES AND NDVI STACK SYNTHESIS

### 2) PlanetScope satellite imagery

Planet Labs is a satellite image provider having constellations of CubeSats which provide 3m resolution PlanetScope imagery and Skysats which provide high-resolution imagery [26]. High resolution data is available commercially. However, for educational program limited PlanetScope data with 3m spatial resolution is available. PlanetScope data is available on daily revisit time. Thus, can be used for high resolution time series. Recently many researchers used Planet data for object detection using machine and deep learning, deforestation studies [27], estimating biomass and crop types, disaster assessments and for mapping buildings and roads using image segmentation technique [28]. Multi-date imagery for three months i.e. for selected areas in Netherlands, April, August and October 2022 and selected area of Pakistan, November 2022, February and March 2023 were acquired in Sentinel-2 harmonized level. This product is normalized to Sentinel 2 bands for consistent radiometry surface reflectance. NDVI was calculated for each of selected month image and temporally stacked to get 3 date NDVI composite image "Fig 4". PlanetScope imagery was also acquired for randomly selected subset areas "Fig 1".

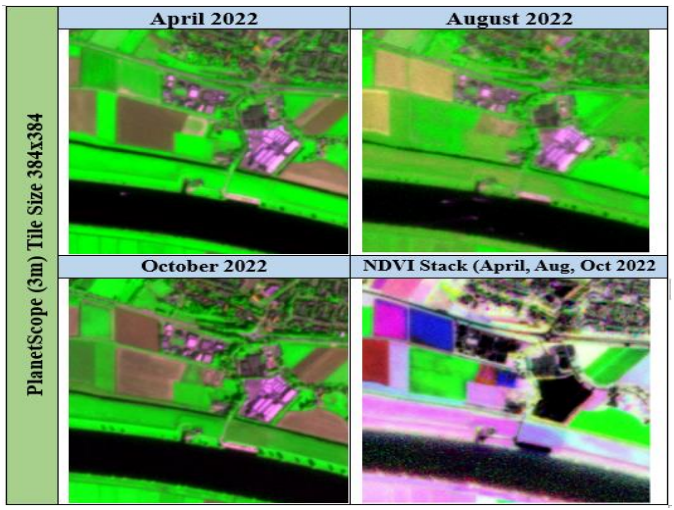

FIGURE 4: PLANETSCOPE, MULTI DATE IMAGES AND NDVI STACK SYNTHESIS

3) Field boundaries ground truth data

For field boundaries delineation a reliable labelled mask for field boundaries is imperative as training of deep learning model. To develop labeled training data containing information about field delimitation and non-field areas, basic registration crop parcels (BRP) vector layer was downloaded online from [29] for Netherlands. It is annual agriculture boundaries layer which is manually digitized and contain information "Fig 5" about crop types, fallow lands, grass lands, non-crop areas and other features.

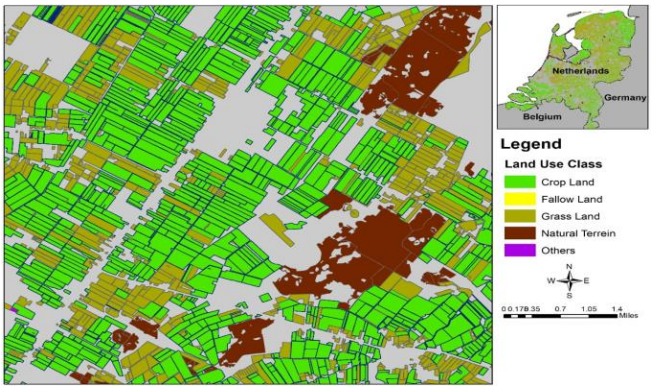

FIGURE 5: SAMPLE AREA SHOWING SIMPLIFIED ATTRIBUTES FOR MAJOR CLASSES IN BRP LAYER OF NETHERLANDS

Due to lack of field boundaries data in Pakistan, labeled vector data for field boundaries was indigenously digitized using visual interpretation on multi-date NDVI composite image and manual annotation (Fig 6).

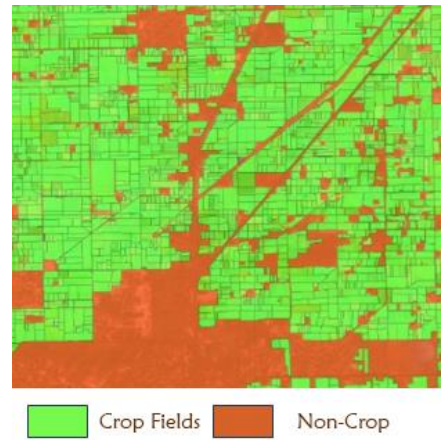

FIGURE 6: SAMPLE AREA SHOWING DIGITISED CROP FIELDS AND NON-CROP AREAS

The BRP layer was clipped for acquired images subset. To convert this layer into labeled mask, crop layer polygons were converted to polyline. A buffer operation was applied keeping in view the native resolution of corresponding imagery i.e 10m for Sentinel-2 and 3m for PlanetScope imagery. These buffered borders were subtracted from original crop polygons to achieve inside field areas and outer boundaries corresponding to resolution of satellite images. Finally, these inside areas and buffered farm outlines were merged and rasterized to get raster labels mask. Final mask had three categories a) 0 for non-crop areas b) 1 for Inner field areas and c) 2 for farm boundaries. Similar procedure was opted for digitized field boundaries vector in Pakistan areas.

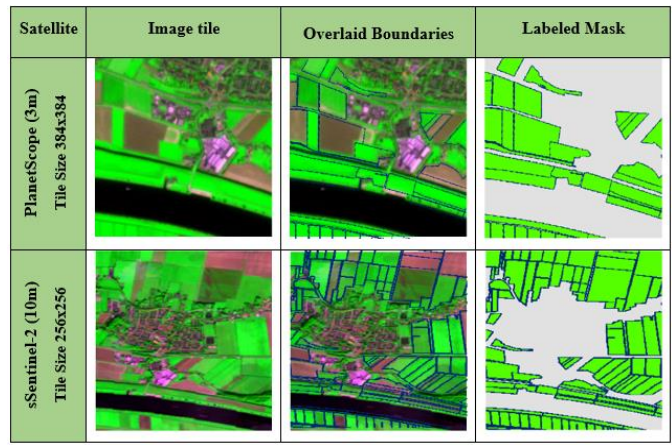

FIGURE 7 : PREPARATION OF LABELLED MASK FOR FIELD BOUNDARIES FROM LPIS VECTOR LAYER

4) Comparative fields size analysis

The "Fig 8" shows histogram comparing the percentage frequency distribution of farm sizes in Pakistan and the Netherlands. The majority of farms in Pakistan are small, with a significant percentage falling in the 0.1 to 0.5 range. Pakistan has a predominance of small farms, whereas the Netherlands has a more diverse range of farm sizes with fewer small farms and more medium to large farms.

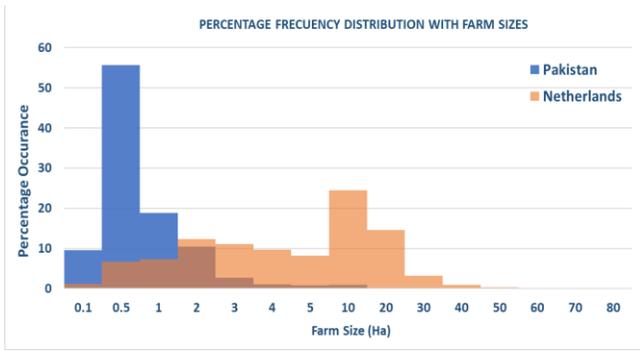

FIGURE 8 : COMPARISON OF FIELD SIZE DISTRIBUTION IN PAKISTAN AND NETHERLANDS

TABLE 1:  MEDIAN FARM SIZE AND NUMBER OF FIELDS IN SELECTED AREAS

|  | Pakistan | Netherlands |
|---|---|---|
| No of Individual farms in selected areas | 6890 | 36826 |
| Median Farm Size (Ha) | 0.34 | 4.21 |

D)  Development of training data for deep learning model

The pre-processing of the satellite data was carried out using open-source image processing libraries in python based platform. The field boundaries in vector form were converted to polylines and buffered to the cell size of the respective raster. The buffered field lines were then rasterized. The final mask subsets were created keeping alignment of original image and mask perfectly and also with the cells of the images. For ingestion to deep learning model, Image tiling was carried out using python based platform.

a)  For Sentinel-2 tile size of 256*256 pixels with 50 percent overlap of 128*128 pixels was adopted.  This produced 1200 tiles for sentinel covering 6.5536 km2 of land area per tile.

b)  For PlanetScope imagery due to higher resolution (3m) than Sentinel-2 (10m) tile size of 384*384 pixels with 50 percent overlap of 128*128 pixels was adopted. This produced 4244 tiles for sentinel covering 1.3271 km2 of land area per tile. The tile size of 256*256 pixel for PlanetScope images results in higher number of sub-tiles while losing contextual information per scene of tile covering less land area. The contextual information caters by using higher tile size but keeping in view compute demand.

c)  For training of models and evaluation, data was split into training, validation and testing @ 70 %, 20% and 10% by randomly selected location grids to minimize the leakage among splits.

### III.    MODELING SETUP

#### A.    Deep learning models implementation

U-Net architecture with four backbones was implemented for this exercise. i) EfficientnetB7 ii) ResNet 50 iii) ResNet 152 and iv) SE-ResNeXt50. Deep residual Network (ResNet) is CNN based model having skip connections between blocks of convolutional and pooling layers. ResNet-152 is deeper network with 230 million parameters [30]. ResNet backbone is one of the popular model with significant accuracy in solving image segmentation problems [31] [32] [27].

To evaluate the usefulness of multidate images for field boundary delineation, 10 combinations of dataset were prepared for training and comparative analysis of Netherlands area (Table 1). Models were trained using Keras/Tensorflow segmentation models library for each set of data using Google Colab. Google Collaboratory or Colab is a cloud-based environment with optional free use which provides GPU resources for limited 12hour runtime. The loss function used for the training was categorical focal dice loss for 3-class segmentations. Model was run for 50 epochs with 10 minibatch size "Table-3".

TABLE 2 SINGLE DATE AND MULTIDATE COMBINATIONS FOR PLANETSCOPE AND SENTINEL-2 DATA

| Satellites | Date | Resolution | Tile Size | Tile Area (km2) | Bands | Total Tiles |
|---|---|---|---|---|---|---|
| PlanetScope | Apr-2022 | 3 m | 384 * 384 | 1.33 | R,G,B, NIR | 4244 |
|  | Aug-2022 | 3 m | 384 * 384 | 1.33 | R,G,B, NIR | 4244 |
|  | Oct-2022 | 3 m | 384 * 384 | 1.33 | R,G,B, NIR | 4244 |
|  | Three Date Bands Stack | 3 m | 384 * 384 | 1.33 | 3R,3G,3B, 3NIR | 4244 |
|  | Three Date NDVI Stack | 3 m | 384 * 384 | 1.33 | NDVI1, NDVI2, NDVI3 | 4244 |
| Sentinel-2 | Apr-2022 | 10 m | 256 * 256 | 6.55 | R,G,B, NIR | 1200 |
|  | Aug-2022 | 10 m | 256 * 256 | 6.55 | R,G,B, NIR | 1200 |
|  | Oct-2022 | 10 m | 256 * 256 | 6.55 | R,G,B, NIR | 1200 |
|  | Three Date Bands Stack | 10 m | 256 * 256 | 6.55 | 3R,3G,3B, 3NIR | 1200 |
|  | Three Date NDVI Stack | 10 m | 256 * 256 | 6.55 | NDVI1, NDVI2, NDVI3 | 1200 |

Multiple deep learning models were evaluated to delineate field boundaries using multi-date NDVI stacks and individual spectral bands. The models tested include EfficientNet B2, ResNet 50, ResNet 152, and SE-ResNeXt 50. The evaluation focused on two different satellite imagery sources: PlanetScope and Sentinel-2, over various dates and combinations. The individual dates included imagery of April, August and October of 2022 for both Planetscope and Sentinel-2. Moreover, models were also trained on composite image of the three dates along with the 3-dates ndvi stack.

Based on the evaluations, the best-performing model and best date of imagery combinations, were selected for further testing and transfer learning to the Dunyapur area in Pakistan. To test the Transfer Learning, the final trained model on the Netherlands dataset was applied to the Dunyapur area to assess its performance on a different geographic region with the same imagery combinations used.

The selected model was further evaluated using the Dunyapur data alone, focusing on the same NDVI stack combinations. Finally, the model was tested on a combined dataset from both the Netherlands and Dunyapur areas to validate its generalizability and effectiveness across diverse agricultural landscapes.

#### B.    Implementation Workflow

Schematic work flow for deep learning implementation of field boundaries delineation is given below (Fig 9)

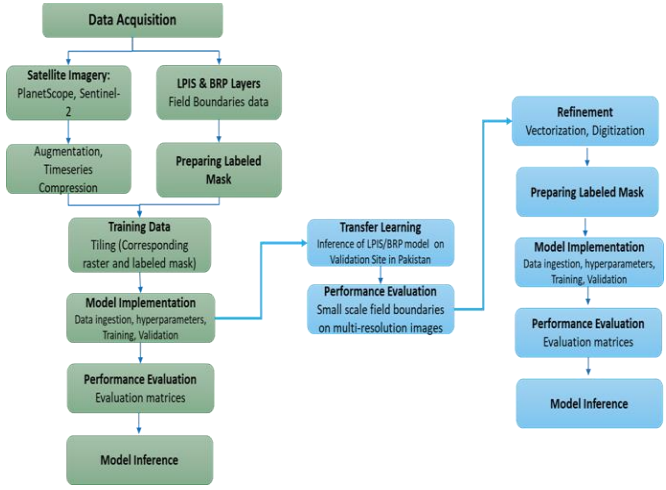

FIGURE 9: WORK FLOW FOR DATA PREPARATION, INGESTION, MODEL TRAINING FOR FIELD BOUNDARY DELINEATION

TABLE 3: MODELS HYPER-PARAMETERS

| Parameter | Value |
|---|---|
| Epochs | 50 |
| Batch Size | 10 |
| Activation | Softmax |
| Optimizer | Adam |
| Loss function | Categorical_Focal_Dice_Loss |
| Model Checkpoint | Monitor Val_Iou_Score |
| Learning Rate | ReduceLROnPlateau on monitoring Val_loss |

### C. Evaluation Metrics

IoU score is a ratio between overlap areas of predictions and ground truth with respect to the union of their areas (Eq-1). Higher IoU value indicates more accurate prediction.

$$IoU = \frac{|A \cap B|}{|A \cup B|} = \frac{|A \cap B|}{|A| + |B| - |A \cap B|} = \frac{TP}{TP + FP + FN}$$

TP = True Positives, FP = False Positives, FN = False Negatives

### D. Post processing and Refinement of predictions

Post-processing enhances the precision of field boundaries by correcting errors from the initial edge detection. Operations like simplify and trim contribute to cleaner, more defined boundaries. This makes the boundaries more practical for subsequent agricultural or GIS applications. In this study the final predictions using best model trained on combined data of Pakistan and Netherlands were post-processed using ESRI's "Geo-processing tool for post-processing of results from edge detection deep learning models". This tool leverages a workflow developed within ESRI's ModelBuilder, allowing for iterative experimentation to optimize results through various geo-processing operations. These operations effectively reduce noise by merging insignificant fragments with significant boundaries.

## IV. RESULTS AND DISCUSSIONS

"FIG 10" presents the mean validation Intersection over Union (IoU) scores of various deep learning models evaluated on different dates and combinations of satellite imagery from the Netherlands. The models tested include EfficientNet-B2, ResNet-50, ResNet-152, and SE-ResNeXt-50. The satellite imagery sources are PlanetScope and Sentinel-2, with the evaluation encompassing different months (April, August, October), composite images, and three-date NDVI stacks.

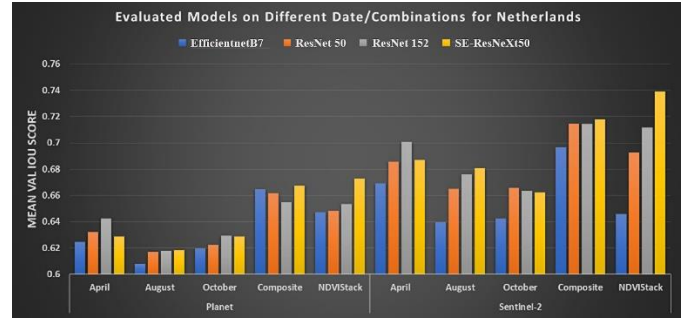

FIGURE 10: COMPARATIVE VISUALIZATION OF IoU SCORES FOR PLANETSCOPE AND SENTINEL-2 IMAGERY

This study was carried under limited compute resource environment. Models were trained with limited number (50) of epochs. The results indicate that the SE-ResNeXt 50 model generally performed better than other models across all dates and combinations, achieving the highest mean IoU scores. ResNet 152 also demonstrates strong performance, particularly with Sentinel-2 data, where it closely follows SE-ResNeXt 50. ResNet 50 shows better results than EfficientNet B2, which, while performing reasonably well, achieves the lowest scores among the four models.

For PlanetScope imagery, the NDVI stack combination yields the highest mean IoU scores, with SE-ResNeXt 50 achieving approximately 0.73. Three dates temporal Composite images also perform well, especially with SE-ResNeXt-50 and ResNet-152. This finding is in-line with [23] where multidate temporal stack improved IoU by 6% in Denmark regions for extraction of field boundaries. However, in this paper, additionally synthesizing the temporal stack in the form of NDVI indices bands not only reduces the input size but also gives similar or better performances than simple multi-date image stacking. Data from Figure 10 shows that IoU scores were less for PlanetScope images than Sentinel-2 images. PlanetScope imagery have better spatial resolution than sentinel-2. It is contrary to findings of [21]. where they observed better F-Score for higher resolution on single date 5m Rapid Eye images than sentinel-2 10m imagery. However, the field sizes in Netherlands are large enough and with enough heterogeneity across fields than sentinel-2 10m resolution appears enough for functional fields discrimination.

Single-date images for April show better performance than those for August and October. In the case of Sentinel-2 imagery, both April images and NDVI stack combinations exhibit better performance. Similar to PlanetScope, composite images and NDVI stack combinations result in higher mean IoU scores. On single date comparisons model performance on August image was less than April and October images both for Planet and Sentinel-2. This explains the phenomenon of dynamic crop development in agricultural fields. On single date, different crops can be at different stage of growth. At early growth stage and due to heterogeneity of crop type static field boundaries may visually more observable than functional crop edges. Coherent fields may also have diffused field boundaries [10]. The SE-ResNeXt 50 model reaches nearly 0.74 with the NDVI stack combination, demonstrating its robustness and effectiveness. This result underscores the benefits of using multi-date NDVI stacks for field boundary delineation in agricultural landscapes across diverse temporal and spectral data combinations.

Keeping in view the effectiveness of multi-date temporal NDVI stacking which gives better functional response for crop fields over space and time but also reduces multi-date input size of images suitable for such studies under limited compute resources, transfer learning was applied using the best-performing models from the Netherlands datasets on NDVI stacks. However, when the trained model of SE-ResNeXt50 3 dates NDVI stack is directly applied to the study area of Dunyapur, there is a sharp decline in prediction performance as shown in "Fig 11". This highlights that the field sizes with respect to the data resolution is a key factor in models being able to delineate field boundaries. Dunyapur region have significantly smaller fields as compared to Netherlands, therefore the model that has only looked at field sizes of Netherlands fails to resolve fields from Dunyapur data.

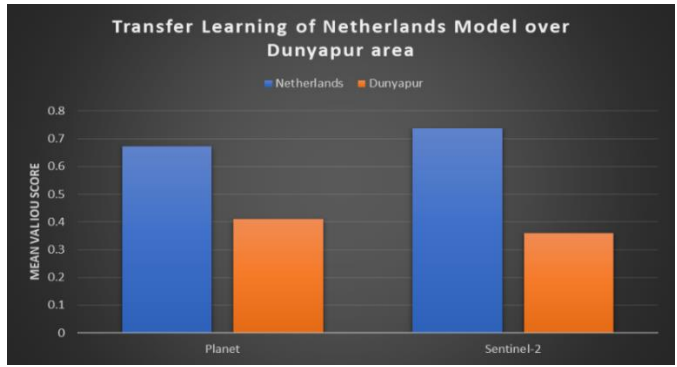

FIGURE 11: TRANSFER LEARNING OF NETHERLANDS MODEL OVER DUNYAPUR AREA

Figure 12 show the mean IoU scores for models trained on i) Three Date NDVI stacks for Netherlands data with BRP layers mask ii) Three Date NDVI stack on Dunyapur images with self-digitized boundaries mask and iii) 3 date NDVI stack combining Dunyapur and Netherlands data together. Result show better mean IoU was achieved on Dunyapur data alone with self-crafted training data than Netherlands data or combined data for both areas. The manually crafted training dataset for Dunyapur visibly correlated with the boundaries in the ndvi stack images whereas the boundaries for Netherlands were marked as parcels. A separate model was trained on this dataset and IoU scores showed a considerable improvement for Dunyapur even more than those of Netherlands. However, predictions from trained models over two geographical areas were not did not generalize well and provide accurate predictions across these different regions. Whereas when model was trained by combining two geographical areas together achieved better mean IoU score (0.74) and also generalized well. This highlights the importance of multi-scale ground information representing different geographical areas for a well generalized or Universal model for field boundaries delineation tasks especially for the areas with small landholding systems and lack of ground data for model training.

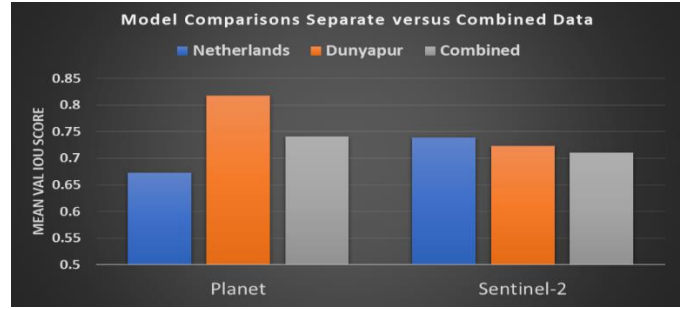

FIGURE 12 MEAN IoU SCORES FOR MODELS TRAINED ON NETHERLANDS, DUNYAPUR AND COMBINED DATA

V. POST-PROCESSING AND REFINEMENT OF PREDICTIONS

Initially, the segmentation results exhibited non-closed and excessive field boundaries, as depicted in Figure 14b. However, the systematic application of the post-processing steps outlined in the methodology using ESRI's tool significantly improved the field shapes and outlines, as shown in Figure 14c. The expansion and subsequent vectorization ensured continuous boundaries, while the simplification and elimination steps removed extraneous features, resulting in a more accurate representation of the field boundaries [33].

| Observations | Image Chips, labels and Predictions |
|---|---|
| a) Better results on large fields on Single Date | |
| b) Diffused prediction in shadowed and dull areas. | |
| c) Useful results on Oct date imagery | |
| d) NDVI stack similar looking fields separated | |

FIGURE 13: SAMPLE OUTPUTS FOR IMAGES AND MODEL PREDICTIONS

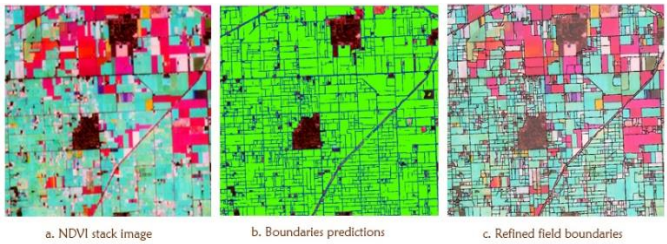

FIGURE 14 FIELD BOUNDARIES PREDICTION AND POSTPROCESSING RESULTS ON DUNYAPUR AREA A. NDVI STACK IMAGE B. MODEL PREDICTION C. POSTPROCESSED REFINED VECTOR OF FIELD BOUNDARIES

## VI. Conclusions

This study successfully highlighted the effectiveness of use of aggregated vegetation index response as multi-date NDVI stack for deep learning model for better field boundaries delineation. Vegetation index derived from different spectral responses i.e. NDVI from Red and NIR bands highlights the vegetation characteristics in single date image. When multi-date NDVIs are stacked they augment additional temporal context in reference to crop growth over different time of season. It may help deep learning model to learn aggregated vegetation response of the crop fields as functional unit and increase the model performance. Visually similar looking field in any single date are separable with NDVI stack.

This study also emphasizes the critical role of multi-scale ground information from diverse geographical areas in developing robust and universally applicable models for field boundary delineation tasks. While predictions from models trained separately on distinct regions showed poor generalization, integrating data from multiple areas significantly improves model performance and generalization. Along the temporal context using multi-date satellite data, there is need of considering geographical diversity and incorporating comprehensive ground data to enhance the accuracy and applicability of predictive models, particularly in regions with fragmented landholding systems and limited ground-truth data. Findings from this study can be extended to multi scale implementation for investigating improved automatic field boundaries delineation in heterogenous agriculture environments.

## ACKNOWLEDGMENT

Special thanks to Planet Labs for providing access to daily PlanetScope imagery through research and academic program.